\documentclass[sigconf, authorversion]{acmart}
\usepackage{bbm}
\usepackage[export]{adjustbox}
{
  \par\vspace{\baselineskip}\noindent
  \textbf{Definition (#1)}\begin{itshape}%
  \par\vspace{\baselineskip}\noindent\ignorespaces
}%
{
  \end{itshape}\ignorespacesafterend
}

\AtBeginDocument{%
  \providecommand\BibTeX{{%
    \normalfont B\kern-0.5em{\scshape i\kern-0.25em b}\kern-0.8em\TeX}}}

\setcopyright{acmcopyright}
\copyrightyear{2021}
\acmYear{2021}
\acmDOI{XXXXXXX.XXXXXXX}

\acmConference[Data Science for Social Networks '21]{Make sure to enter the correct
  conference title from your rights confirmation emai}{December 07--09,
  2021}{Atlanta, GA}
\acmBooktitle{Atlanta '21: Data Science for Social Networks,
December 07--09,
  2021, Atlanta, GA} 
\acmPrice{15.00}
\acmISBN{978-1-4503-XXXX-X/18/06}

\begin{document}

\title{Mitigating Filter Bubbles within Deep Recommender Systems}

\author{Vivek Anand}
\email{vivekanand@gatech.edu}

\affiliation{%
  \institution{Georgia Institute of Technology}
  \city{Atlanta}
  \state{Georgia}
  \country{USA}
}

\author{Matthew Yang}
\email{mattyang@gatech.edu}

\affiliation{%
  \institution{Georgia Institute of Technology}
  \city{Atlanta}
  \state{Georgia}
  \country{USA}
}

\author{Zhanzhan Zhao}
\email{zhanzhan@gatech.edu}

\affiliation{%
  \institution{Georgia Institute of Technology}
  \city{Atlanta}
  \state{Georgia}
  \country{USA}
}

\renewcommand{\shortauthors}{Anand, Yang, and Zhao.}

\begin{abstract}
Recommender systems, which offer personalized suggestions to users, power many of today’s social media, e-commerce and entertainment. However, these systems have been known to intellectually isolate users from a variety of perspectives, or cause filter bubbles. In our work, we characterize and mitigate this filter bubble effect. We do so by classifying various data points based on the user-item interaction history. Subsequently we calculate the influences of these various categories on each other using the well-known TracIn method. Finally, we retrain our recommender system mitigate this filter bubble without compromising accuracy significantly. 
\end{abstract}




\maketitle

\section{Introduction}
Recommender systems, which offer personalized suggestions to users, have become one of the most popular applications of machine learning in today's websites and platforms. 
Prior to the advent of deep learning, recommender systems were generally classified into collaborative filtering (CF) and content based filtering (CB) \cite{parklitreview2012}. For a particular user, CF recommends items that are preferred by other similar users. In contrast, CB uses the similarity between past liked items and future items to recommend items similar to the user's preferences from their data history. 
Nowadays, deep learning based recommender systems significantly outperform classical models \cite{litisasrec2020} \cite{HeNCF2017} and have opened a new chapter for recommender systems due to their capability to both process and fit non-linear data \cite{parklitreview2012}.
However, these recommender systems have been found to be susceptible to detect and amplify preferences of the users or similar users which may lead to \emph{filter bubbles} - the intellectual isolation from a variety of perspectives. Because filter bubbles are especially problematic in social media, due to their potential to cause echo chambers, we use the Twitch dataset \cite{rappaz2021recommendation}. Twitch is an online videostreaming platform where streamers broadcast content to users who can interact with each other and the streamer through the chat. 
With this dataset, we (1) Characterize both training and test datapoints into various categories (2) Calculate the influence of training and test data points between these various categories and (3) Mitigate the filter bubble by retraining with additional data points. With our approach, we are able to increase the diversity of the recommendations by 40-90\% with only minor reductions in accuracy.

\section{Literature Review}
%


Recently, Grosetti et. al \cite{wise2020twitterfilterbubble} quantified how standard recommender systems can affect users' behaviors and amplify filter bubbles with Twitter data. They create profiles for users based on interaction histories, and then generate recommendations for those users using a recommender system. By quantifying diversity of communities with the Gini coefficient measure, they found that 30$\%$ of users receive recommendations that are less diverse than their own user profiles. They label this phenomena as evidence of the filter bubble effect. In our network-based datasets, we can use a similar measurement.


A mitigation strategy is proposed in \cite{wise2020twitterfilterbubble} by re-ranking the outputs from the recommender system algorithms and minimizing an additional objective function---the distance between the user's profile vector and a community score vector generated from the recommendations. The authors were able to mitigate the filter bubble in the Graphjet, CF, and SimGraph recommender algorithms (as defined by Gini coefficients). However, they did not address deep recommender systems and their intervention is post-recommendation time, whereas we focus on intervention at training time and analyze deep models.

Although deep recommender systems dominate in online recommendations, little work has been done to estimate the influence of datapoints on one another. TracIn  \cite{pruthi2020estimating} is one such method to analyze models trained by stochastic gradient descent and its variants, and analysis for regression, text and image classifications are showed in the paper.

\section{Data Description and Analysis}
The dataset we are using consists of interactions on Twitch.tv, provided by Rappaz et. al \cite{rappaz2021recommendation}. Twitch is an interactive livestreaming service for content spanning gaming, entertainment, sports, music, and more. One reason this dataset is a strong fit for our purposes is because interactions on the Twitch platform are two-way, and the streamers can have a large influence on users' beliefs and interests. In addition, the Twitch platform has unique core features such as ``hosting'', in which streamers can send their users to another streamer (thereby contributing to an inherent filter bubble effect). In summary, this dataset has both real-world consequences and is likely to exhibit the filter bubble effect.

The dataset creators used the public Twitch API to query user-streamer interactions over a 43 day period. In their released dataset, each row records a single interaction consisting of: User ID, Stream ID, Streamer, Time start, Time stop. As for raw statistics, the benchmark dataset that we will be modifying contains 100k users, 162.6k streamers, and 3 million interactions. In terms of a recommender system problem, the streamers are the items that are recommended.

Our preprocessing task lies primarily in generating an embedding for each item so that we can use as to measure diversity metric, as well as decreasing the item space so that our model is able to properly train.

\subsection{Community Generation}
As mentioned previously, the diversity of a list of recommendations is typically measured by applying metrics on vector-based item embeddings, or categorical item information. Since our dataset consists of user-item interactions but no actual metadata, we are unable to use standard methods such as Item2Vec \cite{item2vec} to generate vector-based embeddings. Thus, we instead opt to categorize items into communities based off of the network structure of our dataset. Our procedure for community generation is as follows:
\begin{enumerate}
    \item Construct a weighted bipartite graph
    \begin{enumerate}
        \item Each node represents a specific user or item
        \item We denote partite sets $U$ for users and $I$ for items
        \item An edge $e = (u, i, weight)$ exists if $u \in U$ and $i \in I$ interacted in our dataset, with $weight$ equal to the raw number of times this interaction occurs.
    \end{enumerate}
    \item Perform a weighted projection onto the items
    \begin{enumerate}
        \item We project our graph onto the partite set $I$
        \item In the projected graph, an edge $e = (i_{i}, i_{j}, weight)$ exists if $i_{i}$ and $i_{j}$ shared a neighbor $u \in U$ in the original graph, with $weight$ equal to the sum of the shared edges
    \end{enumerate}
    \item Use the Louvain method \cite{louvain} on the projected graph to partition the items into communities$i$
\end{enumerate}

Regarding the Louvain method, the raw number of communities as well as the quality of the communities it produces is heavily dependent on the density of the graph. Since we also want to reduce our item space to achieve high model accuracy, we hyperparameter search over minimum node degree by running the algorithm several times, each time filtering out items with node degree below a certain threshold. See Table ~\ref{tab:community_generation} for the results.



\begin{figure}[h]
  \centering
  \includegraphics[width=0.6\columnwidth]{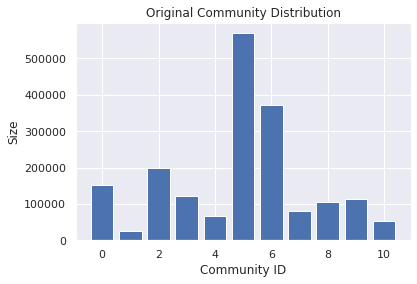}
  \caption{The size of each community in our final partition, weighted by the degree of each item. This is in contrast to partitioning without filtering, in which over half the communities contain a single item.}
  \label{fig:original_community_distribution}
\end{figure}



\begin{table*}
  \caption{Community Statistics by Minimum Item Degree}
  \label{tab:community_generation}
  \begin{tabular}{c|cccc}
    \toprule
    Minimum Item Degree & Number of Items & Modularity & Median Items per Community & Number of Communities\\
    \midrule
    10 & 24546 & 0.5014& 1 & 78 \\
    25 & 11534 & 0.4931 & 632 & 18\\
    50 & 6561 & 0.4836 & 390 & 16\\
    100 & 3735 & 0.4802 & 221 & 13 \\
    250 & 1752 & 0.4616 & 111 & 11 \\
    1000 & 429 & 0.410 & 55 & 8 \\
    \bottomrule
  \end{tabular}
\end{table*}

\subsection{Preprocessing and Sampling}
Now that we have generated a mapping from unique items to community clusters, we can proceed with our preprocessing procedure, which is as follows:
\begin{itemize}
    \item Remove items with <100 interactions (from community generation step)
    \item Remove users with <10 interactions (standard for recommender systems)
    \item Convert item names to integers
    \item Label each user-item interaction with the item's community
    \item Sample a subset of users and take all of their interactions
\end{itemize}

\begin{table}
  \caption{Dataset Statistics: Before and After}
  \label{tab:dataset_statistics}
  \begin{tabular}{ccc}
    \toprule
    Statistic & Original & Final  \\
    \midrule
    Number of users & 100k & 2000 \\
    Number of items  & 162.6k & 1742 \\
    Mean interactions per user & 30.517 & 52.873 \\
    Mean unique interactions per user & 15.052 & 19.062 \\
    \bottomrule
  \end{tabular}
\end{table}
\vspace{-3mm}





\section{Experimental Setting and Baselines}
\subsection{Recommendation Setting}
\textsc{Definition 1: \underline{Next Item Recommendation}} \\
\emph{Given a set of items $I$, let a sequence of items be $S = \{i_{1}, i_{2},... , i_{m}\}$ where $i_{k} \in I$ and $m$ is the length of the sequence. We seek to predict the next $i_{k}$ that given user with a sequence of $\{i_{1}, i_{2}, ..., i_{k-1}\}$ $\forall k \in [1, 50]$ will interact with }.

\subsection{Recommender System Details}
We use a simple Long Short Term Memory (LSTM) \cite{HochreiterLSTM1997} based model as our recommender system. The architecture is as follows.
\begin{itemize}
    \item Embeddings of length 128 for each item in the user-item history (lookback=50) are generated
    \item Embeddings are passed into a single LSTM layer with a hidden dimension of 64.
    \item LSTM outputs are flattened and passed to a fully connected layer with an output size of the number of items (here 1743)
\end{itemize}
We train with a loss function of Categorical Cross Entropy.
\subsection{Dataset and Hyperparameter Details}
For each user, we use the first 80\% of their interactions (temporal) for training. We use the next 10\% of their interactions for validation, and the last 10\% of their interactions for testing.

To train our LSTM based recommender system, we use the Stochastic Gradient Descent Optimizer (SGD) with following hyperparameters. 
\begin{table}[H]
    \centering
    \begin{tabular}{cc}
    \toprule
         Hyperparameter & Value \\
    \midrule
      Batch Size & 2048 \\
      Epochs & 600 \\
      Learning Rate & 5e-3 \\
      Momentum & 0.9 \\
      \bottomrule
    \end{tabular}
    \caption{Hyperparameters used to train the LSTM based recommender system}
    \label{tab:hyperparameters}
\end{table}
\vspace{-8mm}
\subsection{Evaluation Metrics}
With respect to evaluation metrics, observe that we need to evaluate our recommender system on two axis. Firstly, we need to evaluate our recommender system on pure recommendation performance (which is the classical setting). Secondly, we also need to evaluate the diversity of the system's recommendations. Thus, we have a distinct set of metrics for each. During training time we optimize for performance, but the overall goal of this paper is to increase diversity in recommendations. \

\subsubsection{Performance Metrics}
From a recommendation performance standpoint, we use the following metrics.
\begin{itemize}
    \item \textbf{Mean Reciprocal Rank (MRR):} Measures the average reciprocal rank of the ground truth item in the recommendations generated by the recommender system.
    The range of MRR is $\{0, 1\}$. If MRR is closer to 1, then the recommender system is doing better as it gives the ground truth item higher rank. However, if the MRR is closer to 0, then the recommender system is doing worse.
    \item \textbf{Recall@10:} Measures the fraction of times the ground truth item is in the top 10 in the ranked list of items. 
    The range of Recall@10 is $\{0, 1\}$. If Recall@10 is closer to 1, then the recommender system is doing better as the ground truth item is more regularly in the top 10. However, if the Recall@10 is closer to 0, then the recommender system is doing worse.
\end{itemize}

\subsubsection{Diversity Metrics}
From a diversity evaluation standpoint, we use the Gini-Simpson index. Specifically, for a given list of items (historical or recommended), we convert the list to their community embedding. Then, we use the Gini-Simpson index to provide a diversity score.
\begin{itemize}
    \item \textbf{Gini-Simpson Index:} The Gini-Simpson index equals the probability that two items taken at random from a list belong to the same community. We use this diversity metric because it not only measures the types (communities) of items in a list, but also the abundance of each type---heavily punishing lists with uneven distributions. The range of Gini-Simpson is \{0,1\}. If it is closer to 1, then the list of items is more diverse.
\end{itemize}

\subsection{System Information}
We use the DGX station with eight Nvidia Tesla V100-SXM2 GPUs with 32GB RAM each to run our experiments. More details are included with the code at \url{https://github.com/matthewyangcs/mitigating-filter-bubbles-final}.

\subsection{Baselines}

\section{Proposed Method}
Here, we provide our proposed data-centric method for increasing the diversity of the recommendations produced by the LSTM model (which can be applied to any deep recommender system). Specifically, the end-to-end pipeline looks as follows:
\begin{enumerate}
    \item Generate recommendations and analyze the filter bubble
    \item Use the TracIn method to identify influential training data
    \item Augment and/or cleanse the original training data
    \item Train a new model and evaluate the resulting model
\end{enumerate}

\subsection{Filter Bubble Analysis}
In this section, we generate recommendations on the validation dataset and perform analysis to assess whether a filter bubble effect is actually occurring.

Recall that each data point consists of a historical sequence $S = [i_1, i_2, ..., i_{k-1}]$ that we input into our recommender system. Then, we obtain a ranked list of recommendations (items) $Recs = [r_1, r_2,..., r_{50}]$. For each set of experiments, we apply the diversity metric on the recommended items and then compare to the diversity of the historical sequence. Specifically, for each data point of length $m$, we look at the top $m$ recommended items and compare the diversities.

\subsubsection{Relative Diversity of Recommendations} In Grossetti et. al \cite{wise2020twitterfilterbubble}, they quantified the filter bubble phenemonon at a user level. Specifically, they observed that "30\% of [Twitter] users are faced with less diversified recommendations than their own profile", which they used as evidence of the filter bubble effect. 

\begin{figure}[h]
  \includegraphics[width=.5\columnwidth]{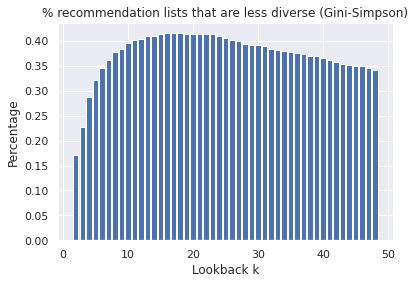}\hfill
  \includegraphics[width=.5\columnwidth]{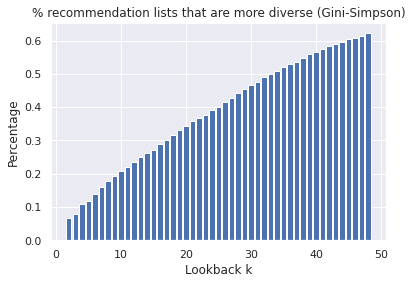}
  \caption{Left: Measures how often the recommendation is less diverse than the history, by various lookback lengths. Right: Measures how often the recommendation is more diverse than the history, by various lookback lengths.}
  \label{fig:relative_diversity}
\end{figure}

\begin{figure}[h]
  \includegraphics[width=0.5\columnwidth]{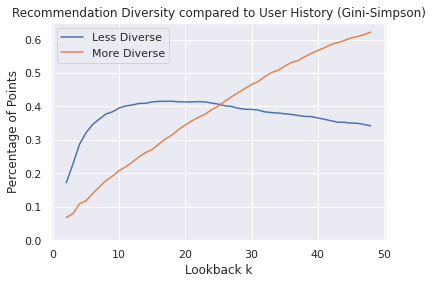}\hfill
\includegraphics[width=0.5\columnwidth]{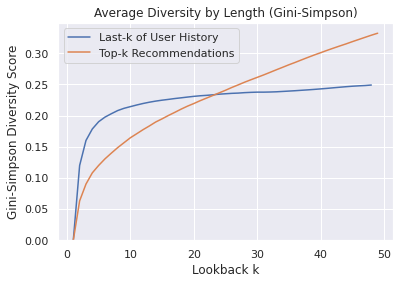}\hfill
  \caption{Left: Measures the relative diversity (by lookback) of the user history and recommendations. Right: Measures the absolute diversity (by lookback) of the user history and the recommendations}
  \label{fig:relative_line_plot}
\end{figure}


Similarly, in this subsection we examine the \% of recommendation lists that are less diverse than the input sequence. The results are reported in Figure ~\ref{fig:relative_diversity}. and ~\ref{fig:relative_line_plot}

\subsubsection{Absolute Diversity of Recommendations} While the filter bubble is technically quantified by the relative diversity of the recommendations, we can also look at the absolute diversity of the recommendations compared to the user history Figure ~\ref{fig:relative_line_plot}).

\subsection{TracIn and Influence Comparison}\label{Influence Comparison}
\subsubsection{Partitioning Training and Validation Data} Because we do not have the computational resources to use Tracin on every single training point, we have to come up with heuristic ways to attempt to capture groups of items that are most influential. Thus, we came up with the following definitions (that we partition our training and validation data into):
\begin{itemize}
    \item Diverse Training Point: A training point that has a history in the top 12.5\% for Gini-Simpson Index
    \item Filter Bubble Training Point: A training point that has a history in the bottom 12.5\% for diversity, and the next item belongs to the majority community in its history
    \item Breaking Bubble Training Point: A training point that has a history in the bottom 12.5\% for diversity, but the next item does not belong to a previously seen community
\end{itemize}

To categorize the predictions, we look at the top-10 predictions on each validation data point. Then, we combine the validation history with each of the top-10 predictions and categorize them into the following:
\begin{itemize}
    \item Filter Bubble Validation Point: A recommendation point that has a history in the bottom 12.5\% for diversity, and the recommended item belongs to the majority community in its history
    \item Breaking Bubble Validation Point: A recommendation point that has a history in the bottom 12.5\% for diversity, but the next item does not belong to a previously seen community
\end{itemize}
\subsubsection{TracIn Applications} After constructing the communities for the various items and classifying the training and test datapoints into various categories, we compute the influence of these categories on each other. The influence function we use here is heavily inspired by TracIn \cite{pruthi2020estimating}. 

TracIn computes the influence of one point on another mainly by using the gradients of their respective losses averaged across all of the checkpoints. 

The original TracIn calculates the influence of the point $z'$ on $z$.  
We, however, use a modified batched version of TracIn for scalability as we only need aggregate influence values. Let $\eta_{i}$ indicate the learning rate at checkpoint $i$, $l$ the loss function and $w_{t_{i}}$ the weights of the network at checkpoint $i$.Let $b$ indicate the batch size and let $\overrightarrow{z}$ and $\overrightarrow{z}'$ be the input vectors of length $b$.

Therefore, for a batch of size $b$ the corresponding influence score will be as follows.
\begin{equation} 
    \text{TracInBatched} (\overrightarrow{z}, \overrightarrow{z}') = \frac{1}{b} \eta_{i} \nabla l (w_{t_{i}}, \overrightarrow{z}) . \nabla l (w_{t_{i}}, \overrightarrow{z}')
\end{equation}

We then use TracInBatched to calculate the influences of the various categories with each other.

\subsection{Data Modification and Re-training}
Steps (3) and (4) in our proposed methods are dependent on the results of 5.2. The idea is that if TracIn is able to identify subsets of our training data that is especially influential on recommending items that exhibit the filter bubble effect, then we can remove those training data points and train a new model, which hopefully will produce more diverse recommendations. Similarly, we can duplicate training data points that have high influence on recommending diversity-increasing data points.

\section{Experiments and Results}
\subsection{Self Influence vs Random Influence}
\label{sec:siri}
We examine the influence of a point on itself, or \emph{self-influence} to better understand the importance of it during prediction time. 

To do this, we take a random subset of size 3000 from our training set and compare the average influences of the training points on themselves. As a control, we compare the average influences of training points on other random training points. To account for random sampling and different batch samples, we repeat this procedure 20 times. 

To determine if self influence indeed is different than the random influence we perform a 2 Sided Welch's independent t-test \cite{welch1956linear} and report the results in the following table. (Independence can be assumed as we take independent random samples of the subsets each time)

\vspace{-1mm}
\begin{table}[H]
    \centering
    \begin{tabular}{cc}
    \toprule
         Statistic & Value \\
    \midrule
      Average Random Influence & 0.047038 \\
      Average Self Influence & 0.081002 \\
      p-value & 3.455549e-28\\
      \bottomrule
    \end{tabular}
    \caption{Self Influence vs Random Influence Comparison}
    \label{tab:si_ri}
\end{table}
\vspace{-8mm}
As the p-value is much smaller than the significance threshold of 0.01, we can conclude that Self Influence is significantly higher than Random Influence.
\subsection{Cross Category Influence}
\label{sec:cross}
Previously we split our training dataset into the various categories of - breaking, filter, and diverse and have split out validation dataset into the categories of - breaking and filter. 

Now, we seek to characterize the influence of all possible combinations of our training categories on the validation categories.
In addition to the above categories, we additionally add random training points as the "random" category and add random validation points as the "random" category to serve as controls for the other combinations.

Due to computational constraints, instead of evaluating on all pairwise influence computations between the each training and validation category, we take 100 samples from both categories and then compute average influences for all pairwise combinations of these two samples. These influence scores are computed with a batch size of 4096. To mitigate the effect of stochasticity, we repeat this experiment 25 times and report the results below.

\begin{figure}[H]
\centering
    \includegraphics[width=0.8\linewidth, left]{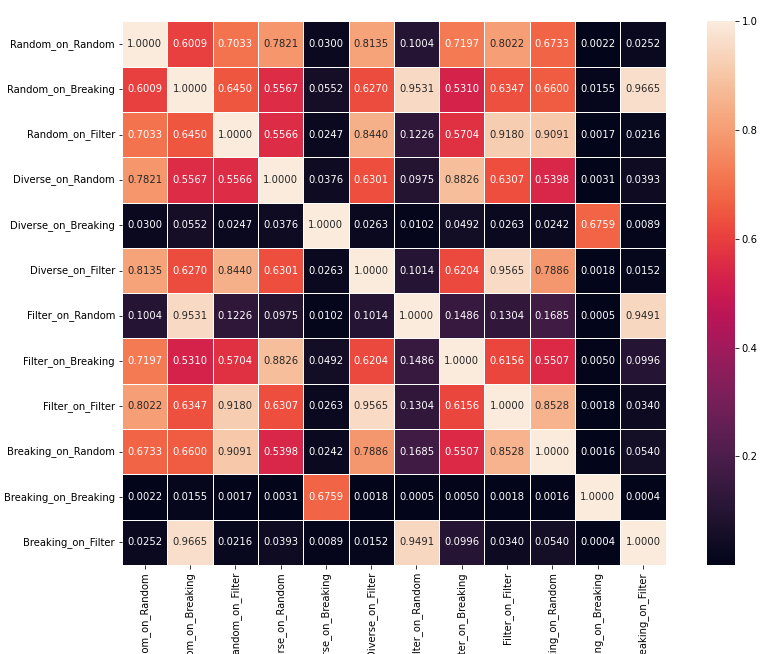}
    \caption{Heatmap showing the p-values comparing the various categories against each other.}
    \label{fig:comparison_heatmap}
\end{figure}

For each category combination, if we assume that the average influence score for each repetition is sampled independently from a distribution, then we can compare the average influences between all the combinations. Figure \ref{fig:comparison_heatmap} visualizes this comparison after conducting Welch's Two Sided Independent t-test \cite{welch1956linear}.

Table \ref{tab:comparison_influence} shows the influence scores of the various training categories on the validation categories

\subsection{TracIn Comparative Analysis}
From Section \ref{sec:siri}, we see that the average self influences is roughly double of the the random influence values. This means that the particular sequence items is more important than a random sequence for reducing the loss of the recommender system on that sequence. This is intuitive and the influence score captures that fact.

From Table \ref{tab:comparison_influence} in Section \ref{sec:cross}, we observe that are only minor effects of the training categories on the validation categories. However, Breaking-On-Breaking and Diverse-On-Breaking are an order of magnitude greater than the other combinations. This indicates that data points that break the filter bubble at training time are critical for reducing the loss of breaking data points in prediction time. Likewise, Diverse points are also critical in reducing the loss of the breaking data in prediction time.  

However, the influence scores are small, about an order of magnitude smaller than the self influence and random influence scores. This is because all of the data points were from the training set, which means that the model is aware that those input data points exist. However, in the cross category influence, the validation data points are not seen at all. Hence, influence is significantly lower.

From Figure \ref{fig:comparison_heatmap}, we see that influence scores for Breaking-On-Breaking, Diverse-On-Breaking and Breaking-On-Filter are statistically different from the other influence scores. However, the other data points are not statistically different. This clearly shows that the type of the data points, in particular the breaking points, in both the training and validation are critical in determining the influence scores.

\begin{table}[H]
    \centering
    \begin{tabular}{ccc}
    \toprule
         Train Category & Validation Category & Influence Score \\
    \midrule
      Random & Random & 0.0046548 \\
      Random & Breaking & -0.0075418 \\
      Random & Filter & 0.0031245\\
      Diverse & Random & 0.0063313 \\
      Diverse & Breaking & 0.052965\\
      Diverse & Filter & 0.0037088 \\
      Filter & Random & -0.0061505 \\
      Filter & Breaking & 0.0076903 \\
      Filter & Filter & 0.0035066 \\
      Breaking & Random & 0.0026899 \\
      Breaking & Breaking & 0.0643030 \\
      Breaking & Filter & -0.0065569 \\
      \bottomrule
    \end{tabular}
    \caption{Influence Scores for Effect of Training Categories on Validation Categories}
    \label{tab:comparison_influence}
\end{table}
\vspace{-7mm}

\vspace{-15pt}
\subsection{Data Augmentation and Cleansing Experiments}
For step (3) and (4) of our proposed method, we use the significant results obtained from the TracIn method to perform training data manipulation experiments. For instance, most notably, we observe that filter bubble training points have high influence on filter bubble predictions, and breaking bubble training points have high influence on breaking bubble predictions. Thus, we propose the following experimental training datasets to evaluate: (1) Remove Filter Bubble Training Points (2) Augment by Duplicating Breaking Bubble Points
(3) Remove Filter Bubble + Duplicate Breaking Bubble.

In addition, we wish to control against the effects of simply removing and duplicating data. Thus, we perform random data removal and augmentation with an equal number of points as the experimental setting (approximately 12.5\%). We obtain the foollowing baselines: (1) Original Training Data (2) Remove a random 12.5\% of the training data (3) Augment by duplicating a random 12.5\% of the training data.

For each experimental setting, we retrain the model 10 times each. Then, we evaluate the performance (MRR, Recall@10) as well as diversity (Gini) (see below). Most notably, this test dataset is separate from the validation dataset, and was unused in any prior experiments. This is important because we based our filtering off of TracIn results on the validation dataset, not the test dataset.

\vspace{-9pt}
\subsection{Data Modification Comparative Analysis}

\vspace{-3mm}
\begin{table}[H]
    \begin{tabular}{cccc}
    \toprule
         Baseline & Recall@10 & MRR & Diversity Index \\
    \midrule
       Original & 0.480 & 0.219 & 0.156 \\
       Remove Random & 0.458 & 0.208 & 0.172 \\
       Add Random & 0.470 & 0.211 & 0.162\\
    \toprule
         Modification & Recall@10 & MRR & Diversity Index \\
    \midrule
      Remove Filter Bubble& 0.465 & 0.213 & 0.176\\
      Add Breaking Bubble & 0.471 & 0.213 & 0.256\\
      Remove and Add & 0.427 & 0.188 & 0.309 \\
      \bottomrule
    \end{tabular}
    \caption{Model performance on validation dataset, using various variations of training data}
    \label{tab:cleansing_results}
\end{table}
\vspace{-8mm}

Table \ref{tab:cleansing_results} Shows the performance under each experimental setting. First, we observe the performance of our model after removing filter bubble points. Compared to using the original training data, we can clearly see that our diversity index is ~12.8\% higher but MRR and recall@10 are lower. But compared to the baseline of removing a random number of training points from the dataset (equal to our number of filter bubble points), we can see that our performance metrics as well as the diversity index is higher across the board. 

Next, we examine the results of adding breaking bubble points. Compared to the original training data, we can see that our diversity index is 64.2\% higher. Compared to a baseline of duplicating a random \% of training points (equal to the number of breaking bubble points we added), we can see that our performance metrics are higher across the board. Thus, we hypothesize that the minor drop in performance compared to the original data is simply because we have a less representative dataset. Further work would involve trying additional gradient methods to identify which breaking bubble points to duplicate.

Lastly, we examine the results of both removing breaking bubble points and adding filter bubble points. This time, we observe the greatest increase in our diversity metric, with a ~98.1\% increase. However, we also experience the largest drop in recall@10 and MRR, which is outside the realm of our additional baselines. The reason for this is likely that we have strayed too far from the representation of the dataset as a whole.

In conclusion, we can see significant improvements in diversity metric with all the proposed methods of data modification, but we do incur performance losses as well, which we will have to investigate in future work. As a final remark, it is worth noting that the drop in recall@10 between the experimental settings and the original baseline is at most 0.043—about 493 mis-predicted test points. Even if we replaced these arbitrarily replaced these mis-predicted points with a perfectly diverse (index=1) of recommendations, it would only result in an increase in average Gini-Simpson index of at most .043—far less than our actual increases in diversity.
\vspace{-2mm}

\section{Conclusions}
Though our work is novel, our work still has some limitations.
We use a very simple LSTM based recommender system, that is not representative of the current recommender systems like TiSASRec \cite{litisasrec2020} or NCF \cite{HeNCF2017}. Results could be different for deep models that take user attributes into account.
Moreover, limiting ourselves to the Twitch Dataset alone could skew results.

Natural extensions of our work would be to expand our suite of experiments to other datasets with the potential for filter bubbles like Twitter or Youtube. That way, we would be able to show how the filter bubble effect and the way to mitigate it could vary depending on the dataset. In the future, we also plan to evaluate this filter bubble effect on state-of-the-art deep recommender systems.

\vspace{-5pt}
\section{Contributions}
\textbf{Vivek Anand:} Conducted Literature Review; setup LSTM pipeline from scratch; set LSTM hyperparameters and setup TracIn checkpointing; researched and implemented all TracIn functions; sped up TracIn 20x with batched approximation; ran all TracIn experiments; wrote majority of proposal and final paper; ran statistical tests; generated all TracIn experiment plots; prepared reproducing TracIn notebooks. \textbf{Matthew Yang:} Conducted literature review and contributed to proposal; researched and developed the procedure for community generation and ran every experiment for community generation and data pre-processing; Wrote vast majority of midpoint presentation; Developed the idea and wrote the code for characterization of training/validation data points for TracIn experiments; Wrote all code for filter bubble exploration from scratch; Wrote and ran all data modification experiments; Produced all plots for my part of the work; Along with Vivek, wrote the entire final presentation and almost the entire final paper. Prepared reproducing notebooks. \textbf{Zhanzhan Zhao:} conducted literature review, validated the training and testing performance of the LSTM model for midterm, and showed the filter bubble effect in (later further modified by Matt).

\vspace{-5pt}
\bibliographystyle{unsrt}
\bibliography{main}

\begin{thebibliography}{10}

\bibitem{parklitreview2012}
Deuk~Hee Park, Hyea~Kyeong Kim, Il~Young Choi, and Jae~Kyeong Kim.
\newblock A literature review and classification of recommender systems
  research.
\newblock {\em Expert Syst. Appl.}, 39(11):10059–10072, sep 2012.

\bibitem{litisasrec2020}
Jiacheng Li, Yujie Wang, and Julian McAuley.
\newblock Time interval aware self-attention for sequential recommendation.
\newblock In {\em Proceedings of the 13th International Conference on Web
  Search and Data Mining}, WSDM '20, page 322–330, New York, NY, USA, 2020.
  Association for Computing Machinery.

\bibitem{HeNCF2017}
Xiangnan He, Lizi Liao, Hanwang Zhang, Liqiang Nie, Xia Hu, and Tat-Seng Chua.
\newblock Neural collaborative filtering.
\newblock In {\em Proceedings of the 26th International Conference on World
  Wide Web}, WWW '17, page 173–182, Republic and Canton of Geneva, CHE, 2017.
  International World Wide Web Conferences Steering Committee.

\bibitem{rappaz2021recommendation}
J{\'e}r{\'e}mie Rappaz, Julian McAuley, and Karl Aberer.
\newblock Recommendation on live-streaming platforms: Dynamic availability and
  repeat consumption.
\newblock In {\em Fifteenth ACM Conference on Recommender Systems}, pages
  390--399, 2021.

\bibitem{wise2020twitterfilterbubble}
Quentin Grossetti, C{\'e}dric du~Mouza, and Nicolas Travers.
\newblock Community-based recommendations on twitter: Avoiding the filter
  bubble.
\newblock In Reynold Cheng, Nikos Mamoulis, Yizhou Sun, and Xin Huang, editors,
  {\em Web Information Systems Engineering -- WISE 2019}, pages 212--227, Cham,
  2019. Springer International Publishing.

\bibitem{pruthi2020estimating}
Garima Pruthi, Frederick Liu, Mukund Sundararajan, and Satyen Kale.
\newblock Estimating training data influence by tracing gradient descent.
\newblock {\em arXiv preprint arXiv:2002.08484}, 2020.

\bibitem{item2vec}
Oren Barkan and Noam Koenigstein.
\newblock Item2vec: Neural item embedding for collaborative filtering.
\newblock In {\em 2016 IEEE 26th International Workshop on Machine Learning for
  Signal Processing (MLSP)}, pages 1--6, 2016.

\bibitem{louvain}
Vincent~D Blondel, Jean-Loup Guillaume, Renaud Lambiotte, and Etienne Lefebvre.
\newblock Fast unfolding of communities in large networks.
\newblock {\em Journal of Statistical Mechanics: Theory and Experiment},
  2008(10):P10008, oct 2008.

\bibitem{HochreiterLSTM1997}
Sepp Hochreiter and J\"{u}rgen Schmidhuber.
\newblock Long short-term memory.
\newblock {\em Neural Comput.}, 9(8):1735–1780, nov 1997.

\bibitem{welch1956linear}
BL~Welch.
\newblock On linear combinations of several variances.
\newblock {\em Journal of the American Statistical Association},
  51(273):132--148, 1956.

\end{thebibliography}

\vspace{-3pt}
\begin{acks}
A special thank you to Sejoon Oh, for providing code for the LSTM recommender system model.
\end{acks}

\end{document}